%% file: main.tex
\documentclass[letterpaper, 10 pt, conference]{ieeeconf}  % Comment this line out if 

\usepackage{graphicx}

\usepackage{todonotes}
\usepackage{booktabs}
\usepackage{lipsum}
\usepackage{subfig}
\usepackage{mciteplus}
\usepackage{url}
\usepackage{amssymb}
\usepackage{pifont}% http://ctan.org/pkg/pifont
\usepackage{svg}
\usepackage{booktabs}
\usepackage[overload]{empheq}

\usepackage[font=small]{caption}


\input{00_defines.tex}

\title{\LARGE \bf Gimme' Signals: Discriminative signal encoding for multimodal activity recognition}

\IEEEoverridecommandlockouts
\author{Raphael Memmesheimer \and Nick Theisen \and Dietrich Paulus% <-this % stops a space
    \thanks{All authors are with the Active Vision Group, Institute for Computational Visualistics, University of Koblenz-Landau, Germany}%
    \thanks{Corresponding email: raphael@uni-koblenz.de}
}

\begin{document}
\maketitle
\thispagestyle{empty}
\pagestyle{empty}

\begin{abstract}
\input{00_abstract.tex}

\end{abstract}

\section{Introduction}
\input{01_introduction.tex}

\section{Related Work}
\input{02_related_work.tex}

\section{Approach}
\input{03_approach.tex}

\section{Experiments}
\input{04_experiments.tex}

\section{Conclusion}
\input{05_conclusion.tex}
\bibliographystyle{IEEEtranM}
\bibliography{references}

\end{document}

%%%%%%%%% TITLE
\title{(Supplementary) Gimme Signals: Discriminative signal encoding for multimodal activity recognition}

\author{Raphael Memmesheimer\\
Institution1\\
Institution1 address\\
{\tt\small firstauthor@i1.org}
\and
Second Author\\
Institution2\\
First line of institution2 address\\
{\tt\small secondauthor@i2.org}
}

\maketitle

\section*{Content}

\begin{figure*}[t!]
    \centering
    \includegraphics[width=\linewidth]{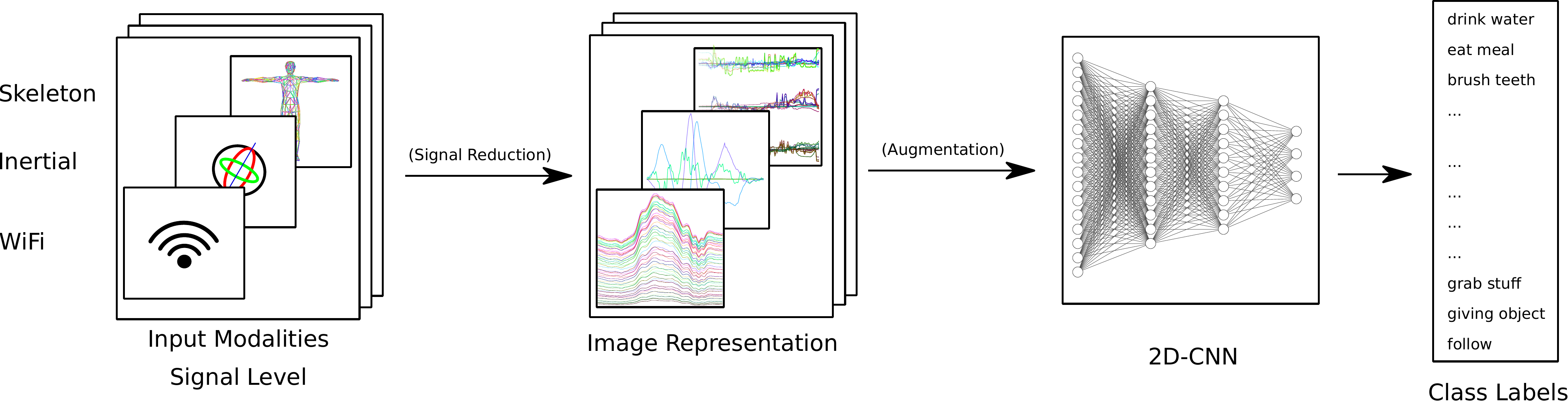}
    % \includesvg[width=\linewidth]{images/overview_2.svg}
    \caption{Approach overview.}
    \label{fig:approach_overview}
\end{figure*}

\begin{figure*}[t!]
    \centering
\includegraphics[width=\linewidth]{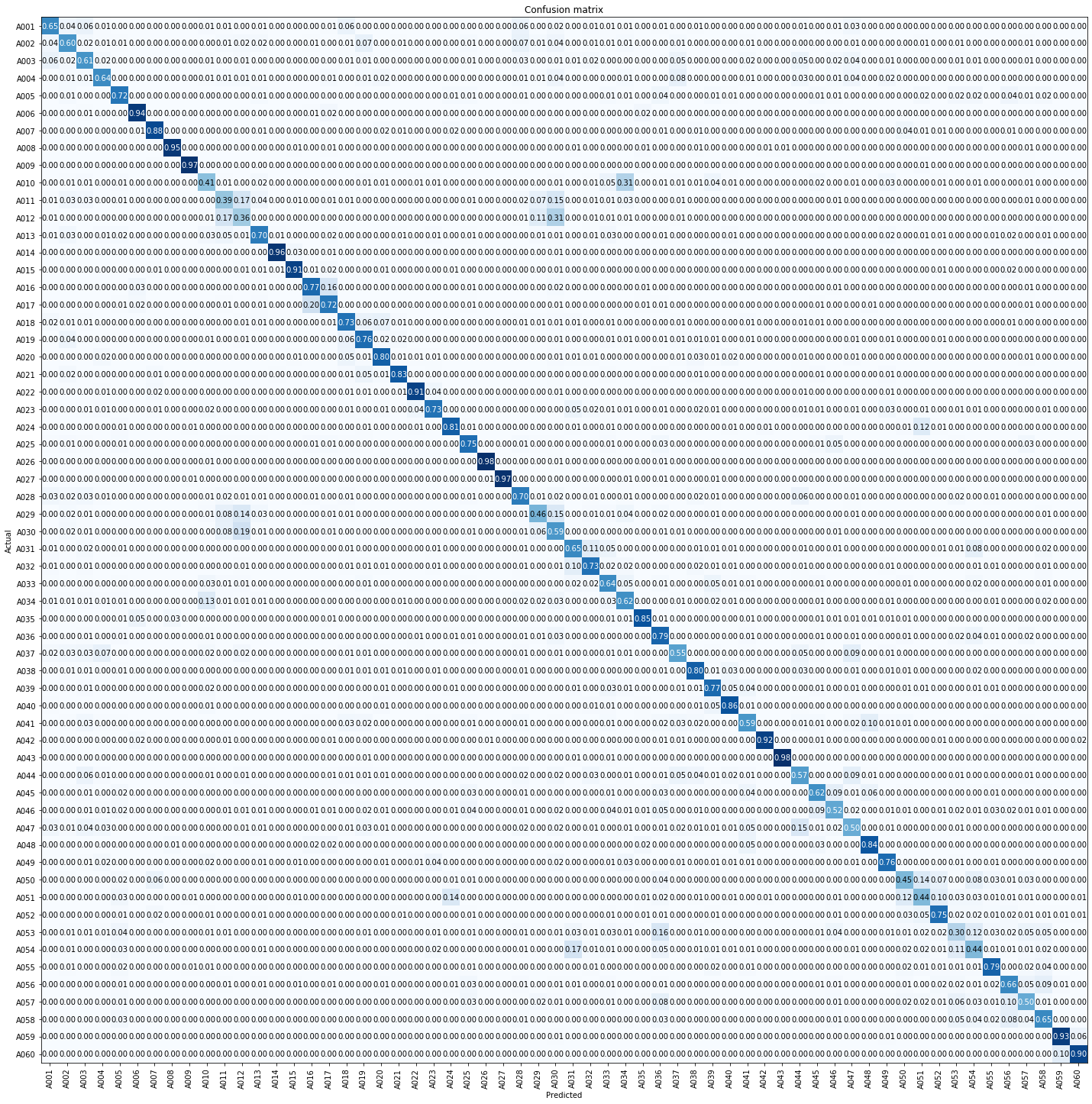}
\caption{Confusion Matrix for NTU-60 (AIS)}
\label{fig:confusion_ntu_60_ais}
\end{figure*}

\begin{itemize}
    \item Figure \ref{fig:approach_overview} gives an additional overview of the proposed approach.
    \item Table \ref{tab:results_ntu} shows results on the NTU-60 subset of the NTU-120 dataset.
    \item Figure \ref{fig:confusion_utd_mhad_skeleton} shows a confusion matrix of our results on the UTD-MHAD dataset (Skeleton + AIS).
    \item Figure \ref{fig:confusion_aril_ais} shows a confusion matrix of our results on the ARIL dataset (AIS).
    \item Figure \ref{fig:confusion_ntu_60_ais} shows a confusion matrix of our results on the NTU 60 dataset (AIS).
    
    \item Results on the NTU-120 Cross View split and Simitate dataset (MoCap) are not attached to not conflict with the \textit{Supplementary Material Submission} guidelines.
\end{itemize}

% \section{Additional Tables}

% \subsubsection{Simitate \cite{memmesheimer2019simitate}} The Simitate benchmark focuses on imitation learning tasks. We integrate this benchmark as we want to imitation by detection methods. Hand and object data is provided from a motion capturing system. Further pre-calculated 2D pose estimates are provided. The accompanied dataset consists of 1932 sequences. The individuals execute 
% tasks of different kinds of activities from drawing motions with their hand over to object interactions and more complex activities like ironing. The motivation behind using this dataset is, when those actions can be correctly recognized and the robot has a set of actions that it can map the observed actions to we can imitate the observed behaviour by detection using different Resnet architecture depths for 500 Epochs. Results are reported in \tabname <2\ref{tab:results_simitate}.

\begin{table}[]
    \centering
\begin{tabular}{l|r|r}
\textbf{Approach}     & \textbf{Accuracy} & \textbf{Top-5 Accuracy} \\
\toprule
Ours (Raw, Resnet152)   & 0.943470 & 0.988304 \\
Ours (AIS, Resnet152)   & 0.961014 & 0.992203 \\
% Ours (Raw, Resnet101)   & 0. & 0. \\
% Ours (AIS, Resnet101)   & 0. & 0. \\
Ours (Raw, Resnet50)   &  0.925926 & 0.984405\\
Ours (AIS, Resnet50)   & 0.959064  & 0.994152 \\
Ours (Raw, Resnet18)   & 0.925926 & 0.984405 \\
Ours (AIS, Resnet18)   & 0.966862 & 0.996101  \\
\bottomrule
\end{tabular}
    \caption{Results on Simitate}
    \label{tab:results_simitate}
\end{table}

% \subsection{UTD-MHAD}

% In addition to the paper we report the fused results of inertial + skeleton data in \tabname \ref{tab:results_utd}.

% \begin{table}[]
%     \centering
% \begin{tabular}{l|r}
% \textbf{Approach} & \textbf{Accuracy} \\
% \toprule
% Zhao \andothers \cite{zhao2019bayesian} & \textbf{92.8} \\
% Wang \andothers \cite{wang2018action}  & 85.81\\
% Chen \andothers (Kinect DMMs) \cite{chen2015utd}  & 66.1\\
% Chen \andothers (Inertial) \cite{chen2015utd}  & 67.2\\
% Chen \andothers (Fused) \cite{chen2015utd}  & 79.1\\
% \midrule
% % Signal Skeleton\\
% % \midrule
% Ours (Skeleton) & 81.16 \\
% Ours (Skeleton, AIS) & 92.09\\
% % Ours (ASS) & TODO\\
% % \midrule
% % Signal Inertial\\
% % \midrule
% Ours (Inertial) & 50.23\\
% Ours (Inertial, AIS) & 69.30\\
% Ours (Fused)   & 63.95 \\
% Ours (Fused, AIS)   & 82.56 \\

% % Ours (ASS) & TODO\\
% % line\_5\_seperated\_axis & Skeleton & 0.883721\\
% % line\_5\_seperated\_axis & Inertial & 0.7 \todo{Redo experiment, number lost}\\
% % \midrule
% % 3d representation\\
% % \midrule
% % line\_5 Skeleton & 0.755814 \\
% % dot\_5 Skeleton  & 0.709302 \\ 
% \bottomrule

% \end{tabular}
%     \caption{Results on UTD-MHAD. Units are in\%}
%     \label{tab:results_utd}
% \end{table}

% \subsection{NTU-60}
\begin{figure*}[t!]
    \centering
\includesvg[width=.8\linewidth]{./images/stage2_augment_1000_epochs_UTDMHAD_skeletonapproach_id011_gradient_alpha_size_1_no_attention_filter_0.2__augment_001000_epochs_resnet152.svg}
\caption{Confusion Matrix for UTD-MHAD (Skeleton, AIS)}
\label{fig:confusion_utd_mhad_skeleton}
\end{figure*}

\begin{figure*}[t!]
    \centering
\includesvg[width=.8\linewidth]{./images/stage2_augment_1000_epochs_ARIL__augment_000300_epochs_resnet152.svg}
\caption{Confusion Matrix for ARIL (AIS)}
\label{fig:confusion_aril_ais}
\end{figure*}

\begin{table*}[]
    \centering
    \begin{tabular}{l|r}
        \textbf{Approach}    & \textbf{Cross Subject} \\%& \textbf{Cross View}  \\
        \toprule
        Liu \andothers\cite{liu2018recognizing} & 91.7 \\%& 95.26\\
        Liu \andothers\cite{liu2017enhanced} & 80.03 \\%& 87.21 \\
        Caetano \andothers\cite{caetano2019skelemotion} & 76.5 \\%& 84.7\\
        Kim \andothers\cite{kim2017interpretable} & 74.3 \\%& 83.1 \\
        Ours (AIS) & 72.33\\ %& \ourresultslightntucv\\
        Shahroudy \andothers 2 Layer P-LSTM \cite{shahroudy2016ntu} & 62.93\\%70.27%
        Shahroudy \andothers 1 Layer P-LSTM \cite{shahroudy2016ntu} & 62.05\\ %69.40%
        Shahroudy \andothers 2 Layer LSTM \cite{shahroudy2016ntu} & 60.69 \\ %67.29%
        Shahroudy \andothers 1 Layer LSTM \cite{shahroudy2016ntu} & 59.14 \\ %66.81%
        Shahroudy \andothers 2 Layer RNN \cite{shahroudy2016ntu} & 56.29 \\ %64.09%
        Shahroudy \andothers 1 Layer RNN \cite{shahroudy2016ntu} & 56.02 \\%60.24%
        \bottomrule
    \end{tabular}
    \caption{Approach comparison NTU RGB+D 60. Units are in\%}
    \label{tab:results_ntu}
\end{table*}

% As NTU-60 is a subset of NTU-120 but contains more 

% \subsection{NTU-120}

% In addition to the paper we show results on the Cross View split on 100 epochs. We also show results for 200 epochs instead of intermediate results for the Cross Subject split.

\begin{table}[]
    \centering
    \begin{tabular}{l|r|r}
        \textbf{Approach}    & \textbf{Cross Subject} & \textbf{Cross View}  \\
        \toprule
        % Liu \andothers\cite{liu2018recognizing} & 91.7 & 95.26\\
        % Caetano \andothers\cite{caetano2019skelemotion} & 76.5 & 84.7\\
        % Kim \andothers\cite{kim2017interpretable} & 74.3 & 83.1 \\
        % Liu \andothers\cite{liu2017enhanced} & 80.03 & 87.21 \\

        % \textcolor{red}{Ours} & \ourresultslightntucs & \ourresultslightntucv\\
        % \midrule
        % NTU RGB+D 120 \\
        % \midrule
        Shahroudy \andothers \cite{shahroudy2016ntu} & 25.5 & 26.3 \\
        Hu \andothers \cite{hu2018early}       & 36.3 & 44.9 \\
        Hu \andothers \cite{hu2015jointly}& 50.8   & 54.7 \\
        Liu \andothers \cite{liu2016spatio}& 55.7 & 57.9 \\
        Liu \andothers \cite{liu2017skeleton1} & 58.2 & 60.9 \\
        Liu \andothers \cite{liu2017global} & 58.3 & 59.2 \\
        Ke \andothers \cite{ke2017new}& 58.4 & 57.9 \\
        Liu \andothers \cite{liu2019skeleton} & 59.9 & 62.4 \\
        Liu \andothers \cite{liu2017enhanced} & 60.3 & 63.2 \\
        Liu \andothers \cite{liu2017skeleton} & 61.2 & 63.3 \\
        Ke \andothers \cite{ke2018learning} & 62.2 & 61.8 \\
        \textit{Ours (AIS)} & 63.62 & 64.86 \\ 
        Liu \andothers \cite{liu2018recognizing} & 64.6 & 66.9 \\
        Caetano \andothers \cite{caetano2019skelemotion} + \cite{yang2018action}  &  67.7 & 66.9 \\
        \bottomrule
    \end{tabular}
    \caption{Approach comparison NTU RGB+D 120. Units are in\%}
    \label{tab:results_ntu}
\end{table}

{\small
\bibliographystyle{ieee_fullname}
\bibliography{references}
}
% distinguishable

%% file: 00_defines.tex
\def\andothers{et al.\ }
\def\figname{Fig.\,}

\def\tabname{Table\,}
\def\wifi{Wi-Fi\ }

\def\ouresultntucs{70.8}
\def\ouresultntucv{71.59}
\def\ntuimpro{+2.9\% (cross-subject), +4.59\% (cross-view)}

\def\arilimpro{+6.78}
\def\arilimproraw{+3.12}
\def\arilresult{94.91} % epoch 75
\def\arilresultraw{91.25}
\def\arilaugmentimpro{3.66}

\def\utdmhadimuresult{81.63} % epoch 87
\def\utdmhadimuraw{72.86}
\def\utdimuimpro{+14.43}
\def\utdmhadinertialaugmentimpro{8.77}

\def\utdmhadfuseresult{86.53} % epoch 87
\def\utdmhadfuseraw{76.13}
\def\utdmhadfusedaugmentimpro{10.37}

\def\utdmhadskeletonresult{93.33} % epoch 67
\def\utdmhadskeletonraw{91.14}
\def\utdmhadskeletonaugmentimpro{2.19}
\def\utdskeletonimpro{+1.13}

\def\simitateresult{96.11}
\def\simitateresultraw{95.72}

%% file: 00_abstract.tex
We present a simple, yet effective and flexible method for action recognition supporting multiple sensor modalities. Multivariate signal sequences are encoded in an image and are then classified using a recently proposed EfficientNet CNN architecture.
% \todo{simple ist vielleicht die nicht passende Übersetzung? Das hat eher den Touch von einfältig/dumm. Vielleicht clever?}
Our focus was to find an approach that generalizes well across different sensor modalities without specific adaptions while still achieving good results.
% \todo{In contrast to most other ist politisch nicht so geschickt. Lass das lieber weg. Diekt:..... achieving results that are comparable (better?) to other published approaches.}
We apply our method to 4 action recognition datasets containing skeleton sequences, inertial and motion capturing measurements as well as \wifi fingerprints that range up to 120 action classes. 
Our method defines the current best CNN-based approach on the NTU RGB+D 120 dataset, lifts the state of the art on the ARIL \wifi dataset by \arilimpro\%, improves the UTD-MHAD inertial baseline by \utdimuimpro\%, the UTD-MHAD skeleton baseline by \utdskeletonimpro\% and achieves \simitateresult\% on the Simitate motion capturing data (80/20 split).
% \textcolor{red}{Describe Results}
% \todo{mit welchem Resultat? Vielleicht sollte "achievable results" lieber hierhin.}
We further demonstrate experiments on both, modality fusion on a signal level and signal reduction to prevent the representation from overloading.

%% file: 01_introduction.tex
\begin{figure}
    \centering
    % \includegraphics[width=\linewidth]{images/examples.png}
    % \includesvg[width=\linewidth]{images/overview.svg}
    \includegraphics[width=.91\linewidth]{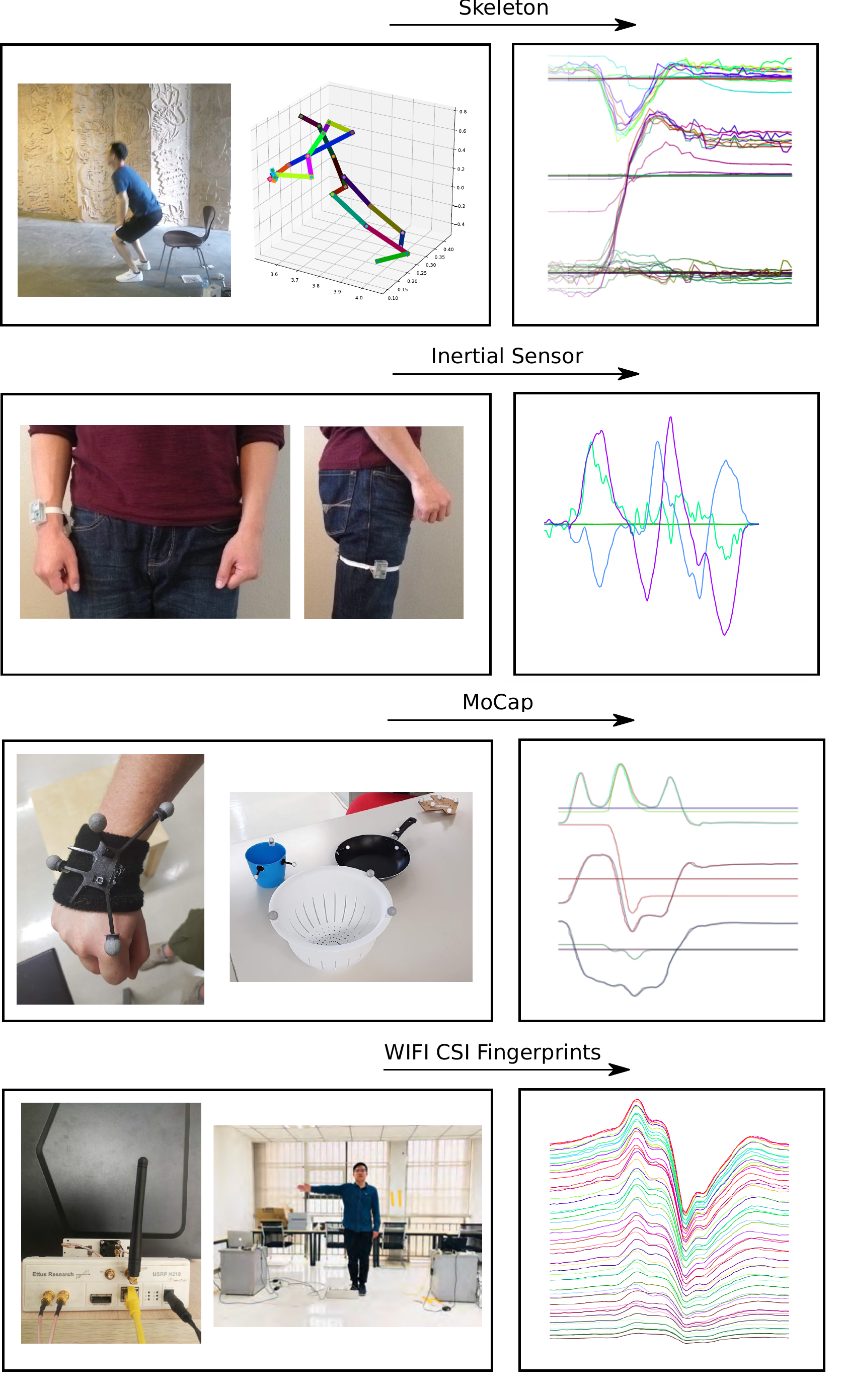}
    \caption{We propose a representation that is suitable for multimodal action recognition.
    The Figure shows representations for skeletal data from the NTU \cite{liu2019ntu,shahroudy2016ntu} dataset,
    Inertial data from the UTD-MHAD \cite{chen2015utd} dataset and WiFi Fingerprints from the ARIL \cite{wang2019joint} dataset. 
    %Our representation is suitable for action recognition on different modalities without specific adaptions.
    }
    \label{fig:overview}
\end{figure}

Action (also referred to as activity or behaviour) recognition is a well studied field and enables application in many different areas like  
elderly care \cite{noury2007fall, solbach2017vision, ni2015elderly, lago2019learning}, 
smart homes \cite{ni2015elderly, lago2019learning}, %\cite{van2010activity, cook2010learning, jalal2012depth, ni2015elderly, van2008accurate, lago2019learning},
surveillance \cite{niu2004human, wiliem2012suspicious} 
robotics \cite{kruger2007meaning, charalampous2016robot} and
driver behaviour analysis \cite{choi2007analysis,martinelli2018human, rigolli2005driver}.

Action recognition can be defined as finding a mapping that assigns a class label to a sequence of signals. The input data can, for instance, be measurements from Inertial Measurement Units (IMU), skeleton sequences, motion capturing sequences or image streams. We tackle the action recognition problem on a signal level as this is a common basis for a variety of input modalities or features that can be transformed into multivariate signal sequences. A common basis is important for the generalization across different modalities.

Some sensors like IMUs, \wifi receivers yield multivariate signals directly, other sensors like RGB-D cameras provide skeleton estimates indirectly. Skeleton estimates can be transformed easily into  multivariate signals by considering their joint axes. This also holds for human poses that can be estimated on camera streams using recent methods \cite{xiao2018simple}.
Predicting the action class from multivariate signal sequences can then be seen as finding discriminative representations for signals. 

Convolutional neural networks have shown great performance in classification tasks. We, therefore, propose a representation that transforms multivariate signal sequences into images. Recent proposed Convolutional Neural Network (CNN) architectures use architecture search conditioned on maximizing the accuracy while minimizing the floating-point operations \cite{sandler2018mobilenetv2, tan2019efficientnet}. Therefore they are good candidates for use in robotic systems.
Figure \ref{fig:overview} gives an exemplary overview of the variety of modalities that our proposed representation can be used for. We evaluated the approach on 5 datasets containing different modalities. 
% To the best of our knowledge, our approach is currently the only one supporting such a variety of modalities without special adaptions of the underlying representation or architecture for each modality. 
Many proposed fusion approaches rely on custom-engineered sub-models per sensor modality which are usually combined in multi-stream architectures. In contrast, we fuse the modalities on a representation level. This has the huge benefit of having a constant computing complexity independent of the number of modalities used whereas multi-stream architectures raise in complexity with every modality added.
% \textcolor{red}{
% While the majority of approaches focus on action recognition on single modalities, there are datasets existing for benchmarking evaluation on multiple modalities \cite{chen2015utd}. Approaches for the multimodal benchmark tend to focus on solving the recognition for each modality separately 

% The 

% We present an approach for action recognition that achieved good results across benchmarks for different modalities.

%}

Our approach lifts the state of the art action recognition accuracy on the ARIL \wifi dataset by \arilimpro\% and the UTD-MHAD \cite{chen2015utd} (IMU \utdimuimpro) (Skeleton \utdskeletonimpro\%). Our approach defines the current best 2D-CNN based approach on the NTU RGB+D 120 dataset (\ntuimpro\%) while being outperformed by a recently proposed graph convolution approach \cite{papadopoulos2019vertex} achieving remarkable results. On the Simitate dataset we achieve \simitateresult\% accuracy on motion capturing data. In total we evaluated our approach on 4 different modalities. To the best of our knowledge, there is no approach showing a comparable high flexibility in supported sensor modalities.

The main contributions of this paper are as follows:
\begin{itemize}
    \item We propose an action recognition approach based on the encoding of signals as images for classification with an efficient 2D-CNN. 
    \item We  propose filter methods on a signal level to remove signals with only a minor contribution to the action. 
    \item We present an approach for information fusion on a signal level.
\end{itemize}
By considering the action recognition problem on a signal level, our approach generalizes well across different sensor modalities. The signal reduction prevents the image representation from overloading and allows flexible addition of signal streams.
By fusion on a signal level, we create a flexible framework for adding additional information for instance object estimates or the fusion of different sensor modalities.
The source code for the presented method is available on github\footnote{\url{http://github.com/airglow/gimme_signals_action_recognition}}.

%% file: 02_related_work.tex
In this section, we present action recognition methods based on traditional feature extractors and recent advances in machine learning. 
Existing survey papers \cite{zhang2016rgb, zhang2019comprehensive, wang2017review, gu2018recent, langkvist2014review} do not include most recent publications as the action recognition field is a highly active field of research. Therefore, most recent approaches from other working groups are presented here.
We put a focus on methods using skeleton sequences as input because these can be acquired on robotic systems directly from RGB-D frames or by extracting human pose features \cite{xiao2018simple} from video sequences.
Further, large scale benchmarks \cite{liu2019ntu} are available for action recognition on skeleton sequences, thus a fair comparison of different approaches can be achieved.
% that allows the reproduction of results and comparison of different method. 

\if 0
\textcolor{red}{We present an approach that generalizes well across sensor modalities such as therefore we introduce related action recognition methods on inertial sensors, \wifi signals and the fusion of different modalities. We also present recent methods on direct action recognition methods on image streams for a more general view.}
\fi

% summarize recent methods for action recognition proposed for direct estimation on video sequences and other sensor modalities like inertial or Wi-Fi.
% Most presented approaches are based on action recognition for skeleton sequences, as they have currently the biggest available datasets with up to 120 action classes. 
% As our approach extracts signals from 
% a diversity of sensor modalities we also present methods for time series classification mostly from different domains than usually found in computer vision approaches. 

% There are already survey papers existing, as the action recognition field is under quite active they can not include most recent development. We therefore also present more recent approaches.
% \todo{Aber das hier ist doch kein survey paper? Besser: 
% :}
% \todo{Dieser Paragraph sollte am Anfang stehen. DAfür der erste Satz raus. Dann: We put a focus on methods using skeleton....}

% However, they can not keep up with the high pace of development of novel methods.
% fast development of more recently presented methods. 
An interesting analysis from a human visual perception point of view has been presented by Johansson \cite{johansson1973visual} in 1973. He found that humans are using 10-12 elements in proximal stimulus 
% \todo{ist proximal stimulus allseits bekannt?} 
to distinguish between human motion patterns \cite{johansson1973visual}. This supports the use of skeletons or pose estimation maps as underlying representations 
% as being discriminative 
for activity recognition approaches from a visual perception perspective \cite{si2019attention}.
% \todo{Den Schluss sehe ich nicht. Weil die menschliche Herangehensweise die beste ist per Definition?}
Recent advances in action recognition developed from hand crafted feature extractors to deep learning approaches like 2D- and 3D-CNNs, while in parallel LSTM based methods also improved results on large scale datasets. More recently, graph convolution approaches showed promising results.

% The following section focuses on related approaches. Related datasets are presented in the experiments section.
% Inspired by the findings of Wang \andothers \cite{wang2018action} who discovered
% the ability of using 

\begin{figure*}[th]
    \centering
    \includegraphics[width=\linewidth]{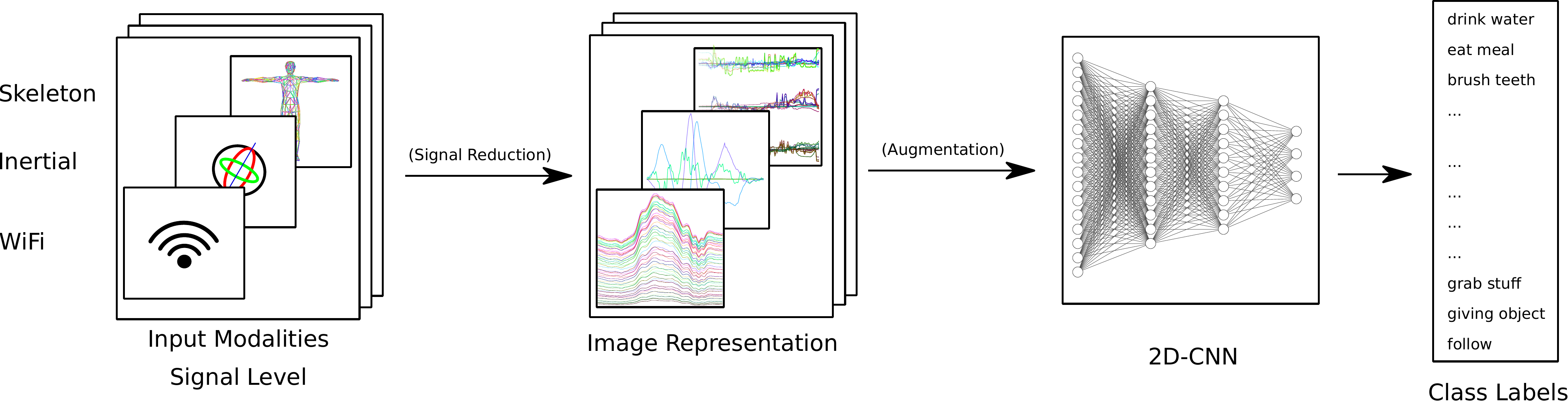}
    % \includesvg[width=\linewidth]{images/overview_2.svg}
    \caption{Approach overview. We propose to transform individual signals of different sensor modalities and represent them as an image. The resulting images are then recognized using a 2D convolutional neural network.}
    \label{fig:approach_overview}
\end{figure*}

% \begin{description}
    % \item 
    % \item[Feature] 
Rahmani \andothers \cite{rahmani2016histogram} presented viewpoint invariant histograms of gradient descriptors for action recognition. Vemulapalli \cite{vemulapalli2014human} represented skeleton joints as points in a Lie-group. The classification is then done by a combination of dynamic time-warping \cite{berndt1994using}, Fourier temporal pyramid representation and linear SVM \cite{vemulapalli2014human}.
    % \item[CNN]
% A lot of research has been investigated in finding representations for 2D-CNNs.
More recent approaches suggest representing skeleton sequences as images and 2D-CNNs for recognition.
%More modern approaches suggested Image representations of skeleton sequences and proposed 2D CNNs for their recognition.
% \todo{nach andothers müssen geschweifte Klammern, sonst kein Leerzeichen...}
Wang \andothers \cite{wang2018action} encode joint trajectory maps into images based on three spatial perspectives. %\todo{spatial? visual? emotional?}
Caetano \andothers \cite{caetano2019skelemotion, caetano2019skeleton} represent a combination of reference joints and a tree-structured skeleton in images. Their approach preserves spatio-temporal relations and joint relevance. Liu \andothers \cite{liu2018recognizing} study a pose map representation. 
The approach that comes closest to our approach is by Liu \andothers \cite{liu2017enhanced}. 
% Kim \andothers \cite{kim2017interpretable}.
Liu \andothers presented a combination of skeleton visualization 
%\todo{US oder GB-englisch? visualisation/visualization, behaviour/behavior} 
methods and jointly trained them on multiple streams. 
% encode signals in a slightly different way by representing the individual axes color encoded. 
In contrast to our approach, their underlying representation enforces custom network architectures and is constrained to skeleton sequences whereas our approach adds flexibility to other sensor modalities.
% does not show generalization across other modalities. 
% For different joints they use the same encoding. This is in contrast to our representation, as we interpret each joint axis individual and encode them respectively. 
% \textcolor{red}{
Kim \andothers \cite{kim2017interpretable} presented a visual interpretable method for action recognition using temporal convolutional networks. Their approach uses a spatio-temporal representation which allows visual analysis to understand why a model predicted an action. Especially joint contributions are visually interpretable.

    % \item[3D CNN]
3D convolutions for video action recognition was popularized by Tran \andothers \cite{tran2015learning}. They have shown good performance on direct video action classification
A three-stream network has then been proposed to integrate multiple cues sequentially via a Markov chain model \cite{zolfaghari2017chained}. By the integration of additional cues from e.g. pose information, optical flow and RGB images using a Markov chain they could increase the recognition accuracy incremental with each additional cue. 
% A special focus on more complex actions was put by Hussein \andothers \cite{hussein2019timeception}. They use multi-scale temporal convolutions and thereby reduce the complexity of 3D-CNN architectures to show increased performance on more complex, longer activities.
    
    % \item[Time Series Classification]
CNN architectures for signal classification have also been studied previously in audio processing \cite{hershey2017cnn}. ResNet 1D-CNN architectures have been used for joint classification and localization of activities in \wifi signals \cite{wang2019joint}. For activity classification on a set of inertial sensors Yang \andothers \cite{yang2015deep} acquire time-series signals and classify the activities using a multi-layer CNN.  

% The presented representation is similar to ours and the one by Liu \andothers \cite{liu2017enhanced}. Our approach differs by showing generalization to different sensor modalities and augmentation techniques for further improvement.
    % \item[LSTM]
    
Liu \andothers \cite{liu2016spatio} presented a spatio-temporal LSTM  inspired by graph-based representation of the human skeleton. They further introduced a novel trust-gating mechanism to overcome noise and occlusion.
Si \andothers \cite{si2019attention} presented an Attention Enhanced Graph Convolutional LSTM Network (AGC-LSTM). They use feature augmentation and a three-layer AGC-LSTM to model discriminative spatial-temporal features and yield very good results on cross-view and cross-subject experiments on skeleton sequences. Very recently Papadopoulos \andothers \cite{papadopoulos2019vertex} proposed two novel modules to improve action recognition based on Spatial Graph Convolutional \cite{yan2018spatial} networks. The Graph Vertex Feature Encoder learns vertex features by encoding skeleton data into a new feature space.  While the Dilated Hierarchical Temporal Convolutional Network introduces new convolutional layers capturing temporal dependencies. Currently their approach is leading on NTU-RGB+D 120 \cite{liu2019ntu} dataset. However, their specialization in skeletal representations does not allow direct adaption on different sensor modalities.
    % \item[Fusion]
    
Interesting fusion approaches have been presented previously. Perez \andothers \cite{perez2019mfas} presented an approach for multi-modal fusion architecture search using RGB, depth and skeleton fusion. Song \andothers \cite{song2018skeleton} extract visual features from different modalities around skeletal joints from RGB and optical flow representations. Whereas those approaches have focused on multiple modalities originating from one device (e.g. Microsoft Kinect) there are also methods for the fusion of sensor data from different devices.
% Fusion across complete different sensor setups have also been shown previously. 
Imran \andothers \cite{imran2019evaluating} propose a three-stream architecture, with different sub-architectures per modality. A 1D-CNN for gyroscopic data, a 2D-CNN for a flow-based image classification and an RNN for skeletal classification. In the end, individual features are fused and a class label is predicted. The fused results are promising and additional modalities improved the results. Additional augmentation by signal filter methods has shown to influence the result positively as well. However, the complexity of the architecture and their sub-architectures require engineering and training overhead and lead to increased run-times by each added modality. This is an issue that we overcome by using a common representation for different modalities. 
% We address a common problem but follow a different viewpoint from a representation perspective.
% By this we can rely on already existing architectures and focus discriminative representations. 
Chen \andothers \cite{chen2015utd} fuse depth information, inertial and demonstrate positive influence. However, they also use two different approaches for each modality. Namely, they use depth motion maps for depth sequences and partitioned temporal windows for signal classification of the gyroscope signals.
Most fusion methods rely on complex individual representations per modality or propose complex multi-stream architectures. In contrast, our approach allows modality fusion using matrix concatenations in a single stream.
However, our approach is limited to data which can be represented as 1D signals over time.
By this, our approach is directly usable for a variety of sensors used in robotics like inertial measurement units, MoCap systems or skeleton sequences and can integrate features extracted from higher dimensional image streams that result e.g. in human pose features \cite{xiao2018simple}.

%% file: 03_approach.tex
% We follow the simple idea of encoding individual signals in a discriminative way into an RGB image. By focusing on signals we can encode a variety of different modalities i.e. human skeleton estimates, \wifi and inertial data into an image which then can be classified using a convolutional neural network \cite{he2016deep}. Similar approaches have been previously presented \cite{liu2017enhanced, wang2018action}. However the mentioned approaches focus on the visual representation of skeletons and therefore lose the ability to generalize over different signal sources. 
% An overview of our approach is given in \ref{fig:approach_overview}. 
% We are not presenting a novel architecture, but use the well established Resnet\todo{EfficientNet} \cite{he2016deep} architecture. Models are trained on a discriminative representation, that remains flexible in the ability to encode a varying amount of signal channels. Of course the amount of signals that can be encoded in an image is limited, therefore we propose a signal reduction method that filters signals that have just a minor contribution to the action. Our proposed representation allows to reduce the problem of sensor data fusion to matrix concatenation.

The problem of action recognition with a given set of $k$ actions $Y = \{0,\ldots,k\}$ can be reformulated as a classification problem where a mapping $f: \mathbb{R}^{N\times M} \rightarrow Y$ must be found that assigns an action label to a given input. The input in our case is a Matrix $\mathbf{S} \in \mathbb{R}^{N\times M}$ where each row vector represents a discrete 1-dimensional signal and each column vector represents a sample of all sensors at one specific time step.

After signal reduction the reduced signal matrix $\mathbf{S}_{focused}$ is transformed to an RGB image $\mathbf{I} \in \{0,\ldots,255\}^{H\times W\times 3}$ by normalizing the signal length $M$ to $W$ and the range of the signals to $H$. The identity of each signal is encoded in the color channel.
An overview of our approach is given in \figname \ref{fig:approach_overview}.

\subsection{Signal reduction}

To avoid cluttering of the signal representation we propose a straightforward method for signal reduction which can be used across different modalities. This allows to lay focus on signals with high information content while removing the ones with low information content.

If for example sequences of skeletons are considered many of the joints are not moving significantly throughout the performance of an action. Intuitively it can be understood that when an action is performed while standing in one place the signal of the leg movement does not contribute much to help in classifying the performed action. From this intuition we developed the assumption that low variance signals do contain less information in the context of action recognition as high variance signals. Therefore we propose to set the signals to zero which are not actively contributing to the action by applying a threshold $\tau$ to the signals standard deviation $\sigma$. In our experiments $\tau$ was defined as 20\% of the maximum value of all signals.

To be more concise we define the decision function $f(\vec{s}_j)$ for the $j$-th signal $\vec{s}_j $ in matrix $\mathbf{S}$ as
\begin{equation}
    f(\vec{s}_j) = \begin{cases}
    1,              & \text{if }  \sigma(\vec{s}_j) \geq \tau \\
    0,              & \text{otherwise}.
   \end{cases}
\end{equation}

When applying this function to each signal in matrix $\mathbf{S}$ we receive a vector $\vec{c} \in \mathbb{R}^N$ which encodes in each element if the corresponding signal contributes to the action. By element-wise multiplication of each column vector of $\mathbf{S}$ with $\vec{c}$ $\mathbf{S}_{focus}$ is received where all signals that do not contribute to the action are set to zero. The signals with low contribution to actions are not removed but set to zero to prevent losing the joint identity (encoded in different colors).

Reducing the signals with low contribution to the action reduces the amount of overlapping signals in the image representation which in turn allows to increase the total number of fused signals. We suggest to apply signal reduction prior to fusion, because different scaling of sensor data can result in the elimination of all signals of a sensor with lower variance as another.

\subsection{Signal fusion}

By our formulation the fusion of signals becomes a matrix concatenation: 

\begin{equation}
    \mathbf{S}_{fused} = (\mathbf{S}_{1}|\mathbf{S}_{2}),
\end{equation}

 where $\mathbf{S}_{fused}$ is the fusion of $\mathbf{S}_1$ and $\mathbf{S}_2$ under the assumption that both matrices have the same amount of rows, where rows represent the sequence length. This can be either achieved by subsampling the higher frequency signals or interpolating the lower frequency signals. An example for sensor fusion is the encoding of multiple identities i.e. from skeletal data with $\mathbf{S}_{fused} = (\mathbf{S}_{id1}|\mathbf{S}_{id2})$, where two identities are fused. Another example is  fusion of two sensor modalities with i.e. $\mathbf{S}_{fused} = (\mathbf{S}_{skeleton}|\mathbf{S}_{inertial})$  or adding interaction context by  $\mathbf{S}_{fused} = (\mathbf{S}_{skeleton}|\mathbf{S}_{objects})$. We therefore created a simple framework to support a wide variety of possible applications.

% \left(\begin{array}{ll}a_1&a_2\\a_3&a_4\end{array}\left|\begin{array}{ll}b_1&b_2\\b_3&b_4\end{array}\right.\right).

\subsection{Representation}

% \todo{color seperation}

\begin{figure*}[th] \centering$
  \vspace{0.06in}
  \begin{array}{ccc}
      \includegraphics[width=.23\linewidth]{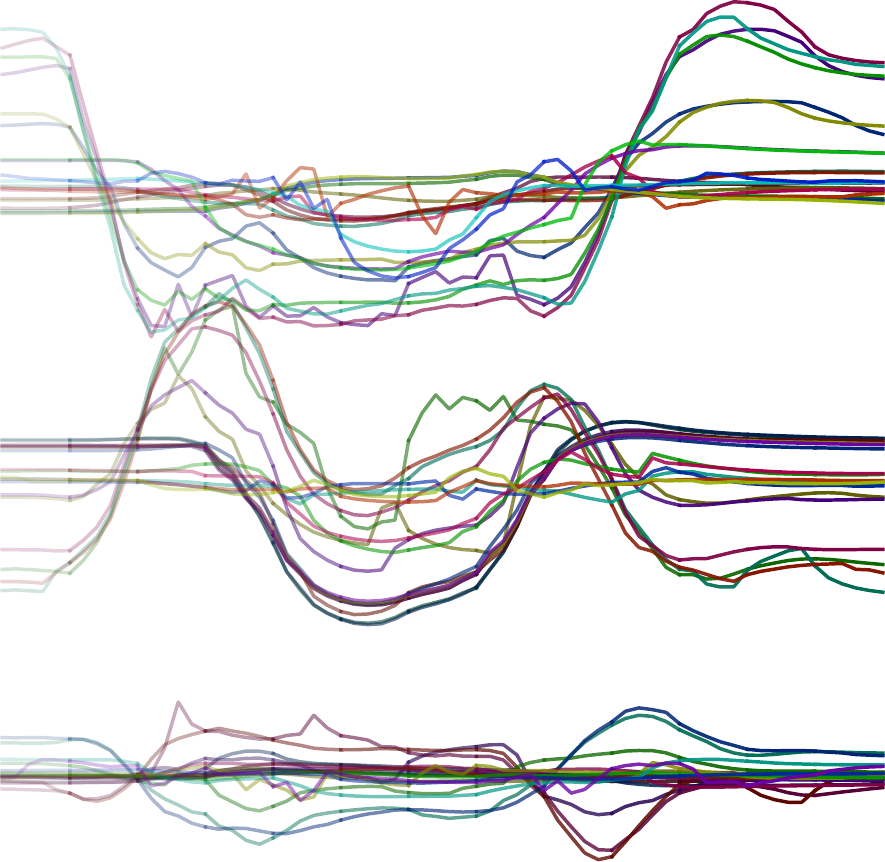} &
      \includegraphics[width=.23\linewidth]{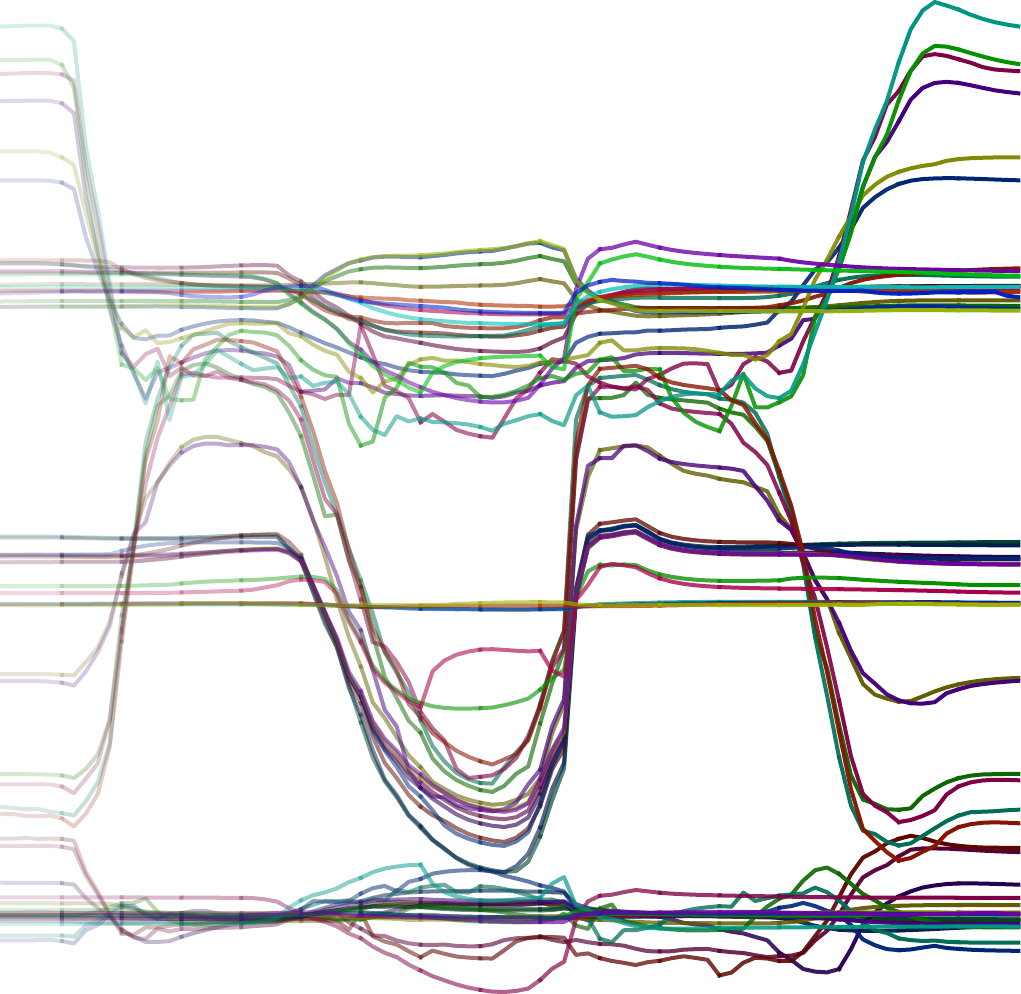} &
      \includegraphics[width=.23\linewidth]{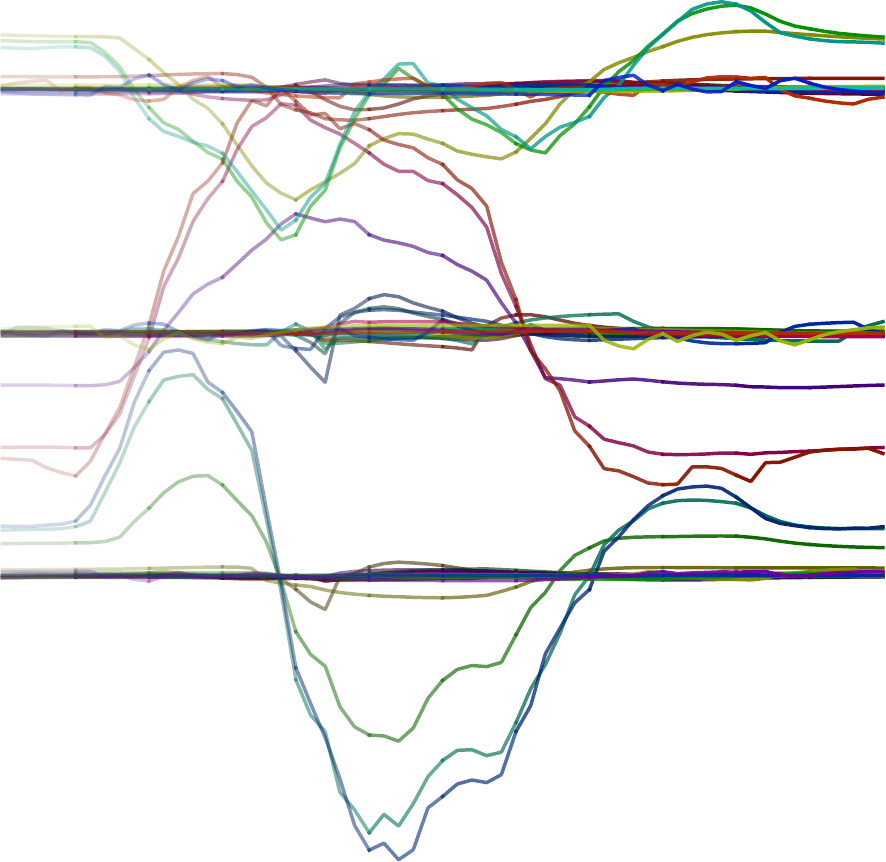} \\
    %   (a) & (b) & (c) \\
      \includegraphics[width=.23\linewidth]{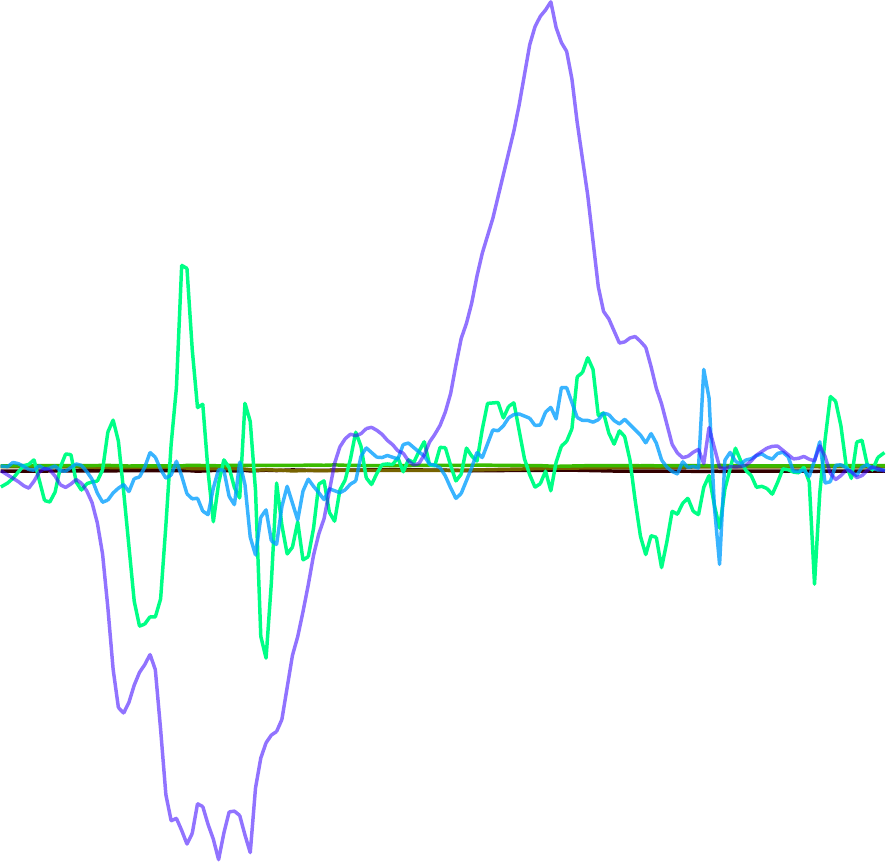} &
      \includegraphics[width=.23\linewidth]{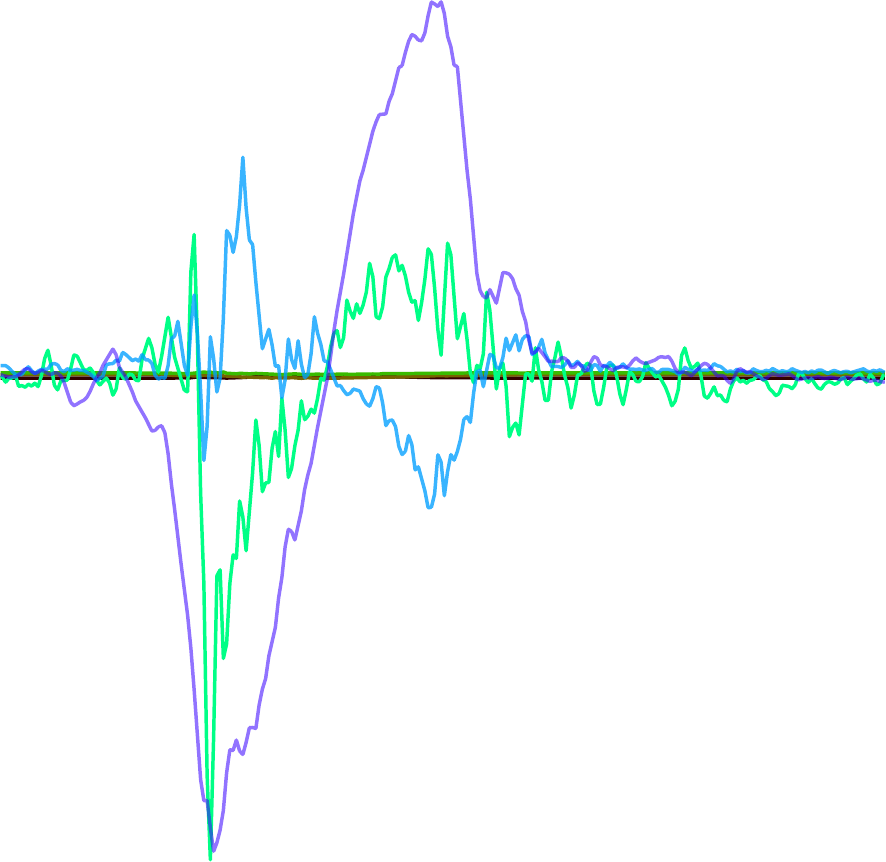} &
      \includegraphics[width=.23\linewidth]{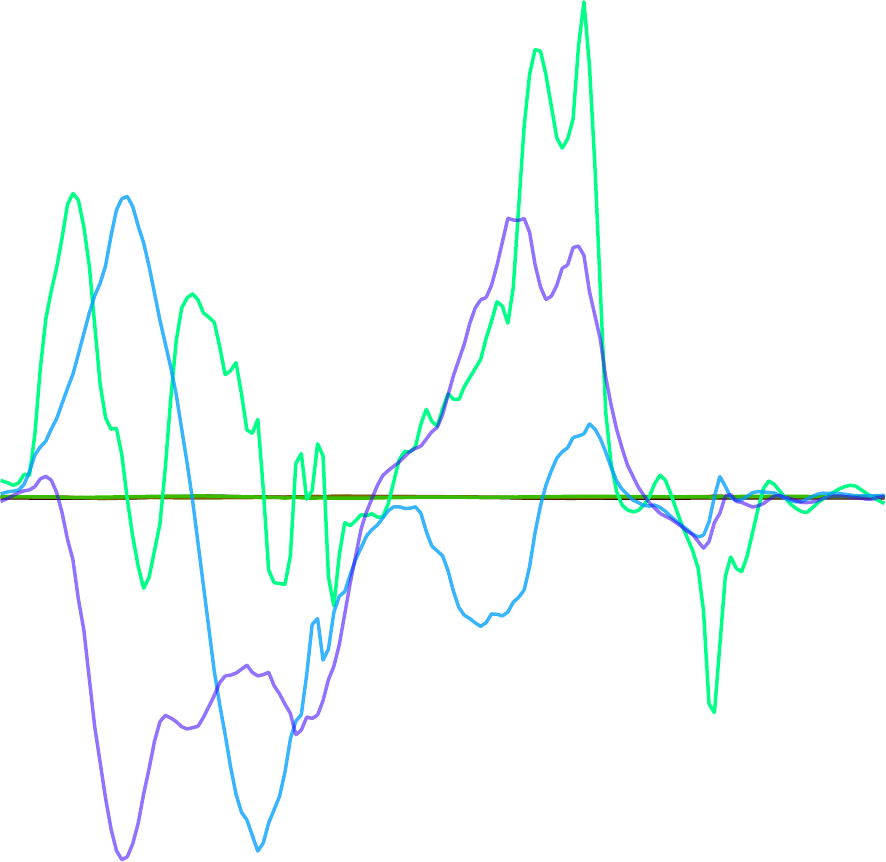} \\
      (a) & (b) & (c)
  \end{array}$
  \caption{Sample representations of the UTD-MHAD dataset: (a) and (b) represent the same class (a27) of different subjects. (c) is a sample of a different class (a1). The color encoded lines correspond to the joint signals. On the top the representation for skeletal data is shown and on the bottom their respective inertial data.}
  \label{fig:utdmhad_examples}
\end{figure*}

\begin{figure*}[th] \centering$
  \vspace{0.06in}
  \begin{array}{ccc}
      \includegraphics[width=.23\linewidth]{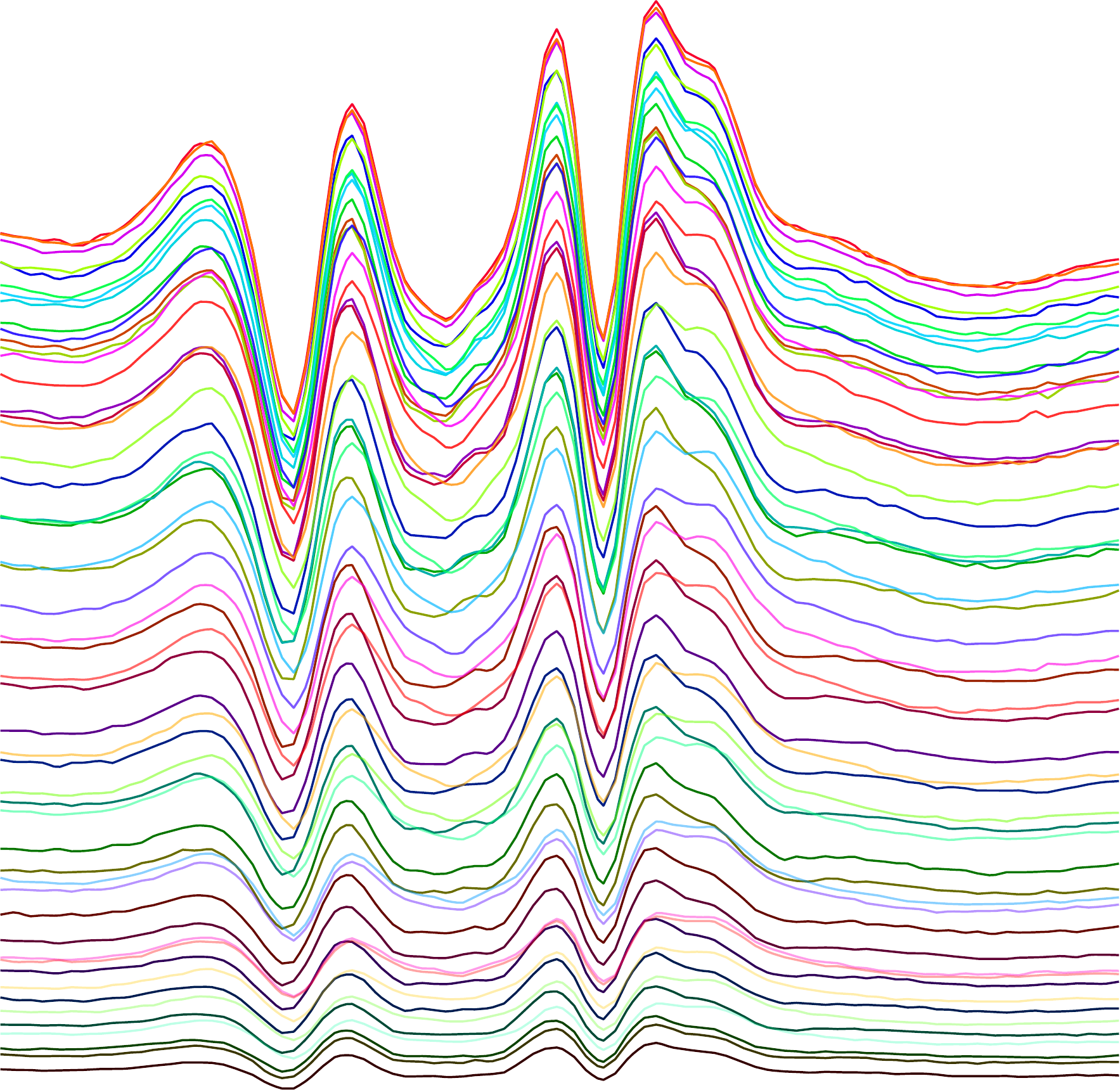} &
      \includegraphics[width=.23\linewidth]{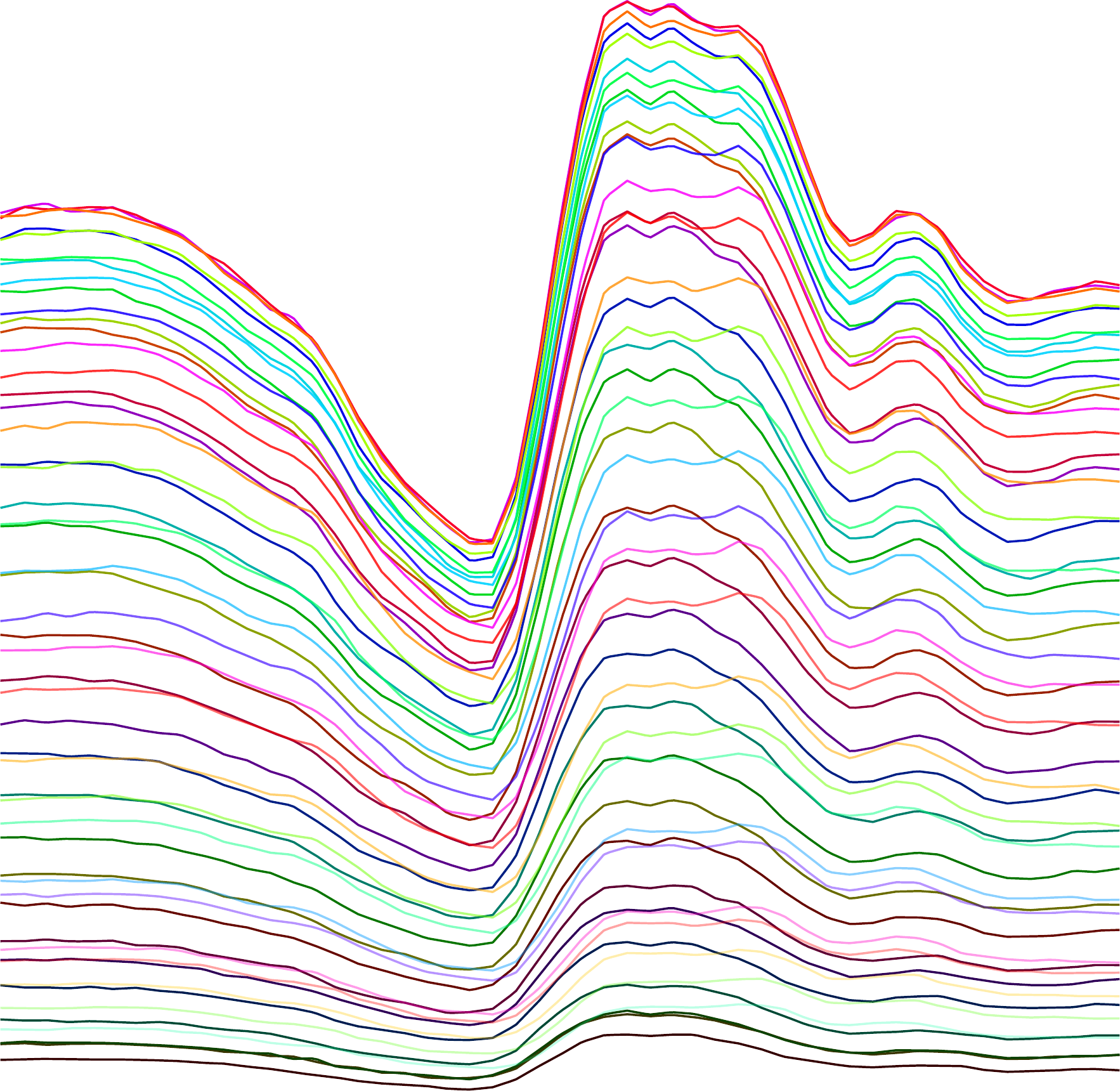} &
      \includegraphics[width=.23\linewidth]{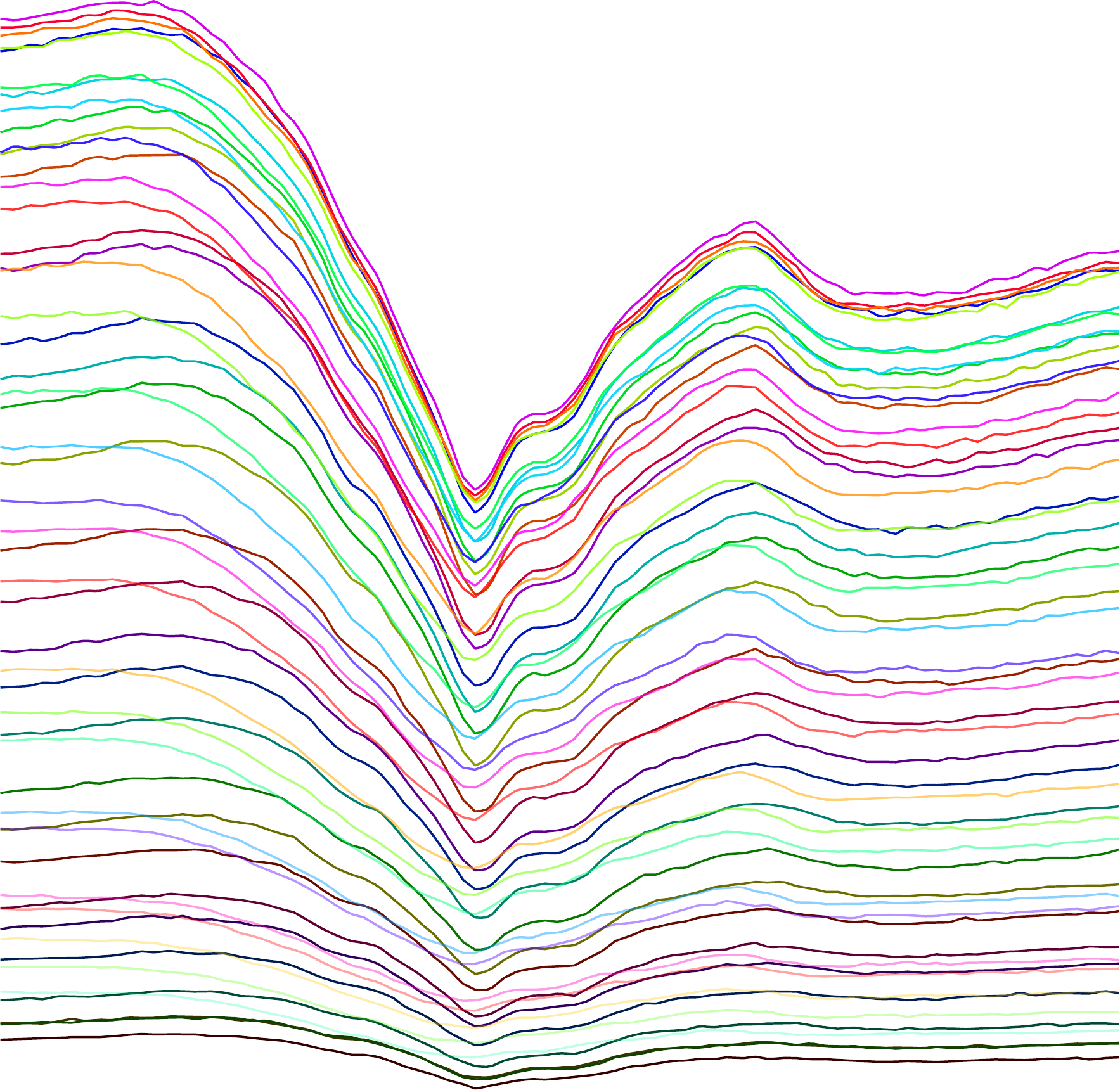} \\
      (a) & (b) & (c) 
  \end{array}$
  \caption{Sample representations: (a) and (b) represent the same class (0) of different subjects. (c) is a sample of a different class. The color encoded lines correspond to the joint signals.}
  \label{fig:aril_examples}
\end{figure*}

To allow a CNN based classifier to discriminate well between the action classes, we aim to find a discriminative representation in the first place. For encoding the signal identity we sample discriminative colors in the HSV color space depending on the number of signals. We make the initial assumption that temporal relations are represented by the position in the image. However, network architectures of lower depth seem to not maintain a global overview of the input but focuses on local relations. Therefore we encode local temporal information by interpolating from white to the sampled color throughout the sequence length. Signal changes are encoded spatially and joint relation are preserved. 
\figname \ref{fig:utdmhad_examples} and \figname \ref{fig:aril_examples} give exemplary representations for skeleton and inertial sequences (\figname \ref{fig:utdmhad_examples}) and \wifi CSI fingerprints (\figname \ref{fig:aril_examples}). A limitation of this approach is that only lower dimensional signals can be encoded. Image sequences or their transformations like optical flow, motion history images are to high dimensional to encode on a signal level by using our representation. Extracted human pose estimates, hand- and/or object estimates from image sequences are adequate signals for encoding in this representation.

\subsection{Augmentation}

Augmentation methods have shown to successfully influence the generalization. 
In our case we can create artificial training data on a signal level by interpolating, sampling, scaling, filtering, adding noise to the individual signals or augment the resulting image representation. Liu \andothers \cite{liu2017enhanced} already proposed to synthesize view independent representations for skeletal motion. As we consider action recognition on a signal level these transformations would result in augmentations integrated as a pre-processing step for each modality separately. Therefore, we decided to focus on augmenting the resulting image representation which can be efficiently integrated into training pipelines. Augmentation applied to the image representation during training still allows interpretation of an effect on the underlying signals. Stretching the width describes the same action but executed in a different speed while perspective changes or rotations can synthesize slightly different executions during the demonstrations.

% Augmentation is now widely used as a generalization technique
% in the classification domain. Most prominently are image based methods \cite{bloice2017augmentor} whereas recently generative approaches have shown improvements for classification \cite{frid2018gan}. Also synthetically generated views \cite{liu2017enhanced} can be interpreted as data augmentation method.
% We experimented with both strategies but found the image space augmentation to be 
% We propose to use augmentation methods on the resulting image representation. 

% Even so the augmentation is executed on the image representation their augmentation can also be interpreted on their effect 
% } 

% \subsection{Training}

% We use the one cycle policy \cite{smith2018disciplined}  for effective training. Further we make use of established recipes \cite{he2019bag} for the training of CNNs. Initialization is done by Kaiming normal \cite{he2016deep}. A learning rate of $0.003$ is used and Adam \cite{kingma2014adam} for optimization. 

\subsection{Architecture}

Most action recognition approaches based on CNNs present custom architecture designs in their pipelines \cite{liu2017enhanced, ke2017new}. A benefit is the direct control over the number of model parameters and can be specifically engineered for data representations or use cases. Recent advances in architecture design can not be transferred directly. Searching good hyper-parameters for training is then often an empirical study. Minor architecture changes can result in a completely different set of hyper-parameters.
He \andothers \cite{he2016deep} suggested the use of residual layers during training resulting in more stable training. Tan \andothers \cite{tan2019efficientnet} recently proposed a novel architecture category based on compound scaling across all dimensions of a CNN. We take advantage of the recent development in architecture design and use an already established architecture for image classification. The recently proposed EfficientNet \cite{tan2019efficientnet} architecture is especially interesting in the robotics context as it's based on architecture search conditioned on maximizing the accuracy while minimizing the floating-point operations.
% Even in the more recent ones no residuals are  

% As we see our contribution in the proposal of a generalizable representation across different modalities we don't propose a new network architecture but focus on the analysis of already established architectures that have proven to be usable for the classification and recognition tasks already. We use
% the widely spread Resnet architecture \cite{he2016deep}. 
% % Especially we use Resnet 18 / Resnet 34 / Resnet 50 architectures to also show the influence of the depth of the architecture. 
% Benefits for using a already existing architecture on the proposed method are the flexible use, the support on reproduction of the results and the integration of additional sensor modalities of different dimensions that have not yet been studied by us.

\subsection{Implementation}

Our implementation is done in Pytorch Lightning \cite{paszke2019pytorch, Falcon2019}, which puts a focus on reproducible research. Hyper-parameters and optimizer states are logged directly into the model checkpoints. The source code is made publicly available. We used a re-implementation and pre-trained weights of the EfficientNet \cite{tan2019efficientnet} architecture. For training we used a Stochastic Gradient Decent optimizer with a learning rate of 0.1 and reduction of learning rate by a factor of 0.1 every 30 epochs with a momentum of 0.9. The learning rate reduction was inspired by He \andothers \cite{he2016deep}. A batch size of 40 was used on a single Nvidia GeForce RTX 2080 TI with 11GB GDDR-6 memory. We trained for a minimum of 150 epochs and used an early stopping policy based on the accuracy after. Similar model checkpoints were created on an increased validation accuracy. For optimizing the training we used a mixed precision approach by training using 16bit float with a 32bit float batch-norm and master weights. A gradient clipping of 0.5 prevented gradient and loss overflows.

% \textcolor{red}{\lipsum[1]}

%% file: 04_experiments.tex
We conducted experiments on 4 different datasets. The NTU RGB+D 120 \cite{liu2019ntu}, UTD-MHAD \cite{chen2015utd}, ARIL \cite{wang2019joint} and the Simitate \cite{memmesheimer2019simitate} dataset. These datasets contain in total 5 modalities.
Skeleton sequences are evaluated on %the NTU RGB+D 60 \cite{shahroudy2016ntu}, 
the recently released NTU RGB+D 120 \cite{liu2019ntu} and the UTD-MHAD dataset \cite{chen2015utd}. The NTU RGB+D 120 dataset demonstrates the scaling capabilities of our approach as it contains 120 classes in more than 114000 sequences. The UTD-MHAD dataset \cite{chen2015utd} provides 27 classes but includes IMU data beside the skeleton estimates. Therefore it is suitable to demonstrate the cross modal capabilities of our approach. We further use it for our fusion experiments. We extend these experiments with activity recognition dataset containing \wifi CSI fingerprints \cite{wang2019joint} and Motion Capturing data from the Simitate \cite{memmesheimer2019simitate} dataset. For our experiments we generated the representations of the datasets prior and used an EfficientNet-B2 \cite{tan2019efficientnet} architecture for classification. AIS in the tables denotes the additional augmentation of the training signals in image space. 
Results are compared to other approaches in the next section.

% \begin{figure*}[t] \centering$
%   \vspace{0.06in}
%   \begin{array}{ccc}
%       \includegraphics[height=3.15cm]{images/} &
%       \includegraphics[height=3.15cm]{images/} &
%       \includegraphics[height=3.15cm]{images/} \\

%       (a) & (b) & (c) 
%   \end{array}$
%   \caption{Example data provided by the datasets used for our experiments. This figure should give an insight of the different underlying representation of the used datasets.}
%   \label{fig:setup}
% \end{figure*}

\subsection{Datasets}
\label{sec:datasets}

%We now introduce the datasets on which we performed our experiments on.
In the following the datasets on which the experiments where performed are introduced.

% \begin{figure*}[t] \centering$
%   \vspace{0.06in}
%   \begin{array}{ccc}
%       \includegraphics[width=.3\linewidth]{images/confusion_utdmhad.png} &
%       \includegraphics[width=.3\linewidth]{images/confusion_nturgbd60.png} &
%       \includegraphics[width=.3\linewidth]{images/confusion_nturgbd120.png} \\

%       (a) & (b) & (c) 
%   \end{array}$
%   \caption{Confusion matrices.}
%   \label{fig:result_confusion}
% \end{figure*}

\subsubsection{NTU RGB+D 120} The NTU RGB+D 120 \cite{liu2019ntu} dataset is a large scale action recognition dataset containing RGB+D image streams and skeleton estimates. 
% And the biggest known to us that contains RGB-D data as well as skeleton data. It contains 56880 sequences with 60 classes executed by 40 subjects from 80 different viewpoints, therefore this dataset is suitable for demonstrating large scale robustness. 
The dataset consists of 114,480 sequences containing 120 action classes from 106 subjects in 155 different views. Cross-view and cross-subject splits are defined as protocols. 
For the cross-subject evaluation, the dataset is split into 53 training subjects and 53 testing subjects as reported by the dataset authors \cite{liu2019ntu}. For the cross-setup evaluation, the dataset sequences with odd setup ids are reserved while the remainder is used for training. Resulting in 16 setups used during training and 16 used for testing. Results are given in \tabname \ref{tab:results_ntu} and are discussed in the next section.
% For cross view results and NTU RGB+D 60 comparison we refer to the supplementary material. 
% The NTU datasets are challenging due to their amount of action classes, subjects and setups and are therefore a good candidate to show generalization on those aspects.
% For the  NTU RGB+D 60 \cite{shahroudy2016ntu} paper results are sourced by their respective authors papers

% \begin{table}
%     \centering
% \begin{tabular}{l|r|r}
% \textbf{Approach}  & \textbf{Cross Subject} & \textbf{Cross View} \\%& CS (Top 5)& CV (Top 5)\\
% \toprule
% NTU RGB+D 60\\
% Signal\\
% \midrule
% line\_5\_seperated\_axis & 0.735333 & 0.735333\\
% \midrule
% NTU RGB+D 120\\
% line\_5\_seperated\_axis & 0.735333 & 0.735333\\
% 3d representation\\
% \midrule

% \bottomrule
% \end{tabular}
%     \caption{Results on NTU RGB+D 60/120}
%     \label{tab:results_ntu}
% \end{table}

\begin{table}[]
    \centering
    \begin{tabular}{l|r|r}
        \textbf{Approach}    & \textbf{CS} & \textbf{CV}  \\
        \toprule
        % Liu \andothers\cite{liu2018recognizing} & 91.7 & 95.26\\
        % Caetano \andothers\cite{caetano2019skelemotion} & 76.5 & 84.7\\
        % Kim \andothers\cite{kim2017interpretable} & 74.3 & 83.1 \\
        % Liu \andothers\cite{liu2017enhanced} & 80.03 & 87.21 \\

        % \textcolor{red}{Ours} & \ourresultslightntucs & \ourresultslightntucv\\
        % \midrule
        % NTU RGB+D 120 \\
        % \midrule
        Part Aware LSTM  \cite{shahroudy2016ntu} & 25.5 & 26.3 \\
        Soft RNN  \cite{hu2018early}       & 36.3 & 44.9 \\
        % Dynamic Skeleton  \cite{hu2015jointly}& 50.8   & 54.7 \\
        Spatio-Termoral LSTM  \cite{liu2016spatio}& 55.7 & 57.9 \\
        % Internal Feature Fusion  \cite{liu2017skeleton1} & 58.2 & 60.9 \\
        GCA-LSTM \andothers \cite{liu2017global} & 58.3 & 59.2 \\
        % Multi-Task Learning Network  \cite{ke2017new}& 58.4 & 57.9 \\
        % FSNet  \cite{liu2019skeleton} & 59.9 & 62.4 \\
        Skeleton Visualization (Single Stream)  \cite{liu2017enhanced} & 60.3 & 63.2 \\
        Two-Stream Attention LSTM  \cite{liu2017skeleton} & 61.2 & 63.3 \\
        Multi-Task CNN with RotClips  \cite{ke2018learning} & 62.2 & 61.8 \\
        Body Pose Evolution Map \cite{liu2018recognizing} & 64.6 & 66.9 \\
        SkeleMotion  \cite{caetano2019skelemotion} & 67.7 & 66.9 \\ 
        TSRJI  \cite{caetano2019skeleton} & 67.9 & 62.8 \\ 
        \textit{Ours (AIS)} & \textit{\ouresultntucs} & \textit{\ouresultntucv} \\ 
        ST-GCN + AS-GCN  w/DH-TCN \cite{papadopoulos2019vertex} & \textbf{78.3} & \textbf{79.2} \\ 
        \bottomrule
    \end{tabular}
    \caption{Results on NTU RGB+D 120. Units are in \%}
    \label{tab:results_ntu}
\end{table}

\begin{table}[]
    \centering
\begin{tabular}{l|r}
\textbf{Approach} & \textbf{Accuracy} \\
\toprule
Zhao \andothers \cite{zhao2019bayesian} & {92.8} \\
Wang \andothers \cite{wang2018action}  & 85.81\\
Chen \andothers (Kinect DMMs) \cite{chen2015utd}  & 66.1\\
Chen \andothers (Inertial) \cite{chen2015utd}  & 67.2\\
Chen \andothers (Fused) \cite{chen2015utd}  & 79.1\\
\midrule
% Signal Skeleton\\
\midrule\textit{Ours (Skeleton)} & \textit{\utdmhadskeletonraw} \\
\textit{Ours (Skeleton, AIS)} & \textbf{\utdmhadskeletonresult} \\
% Ours (ASS) & TODO\\
% \midrule
% Signal Inertial\\
% \midrule
\textit{Ours (Inertial)} & \textit{\utdmhadimuraw} \\
\textit{Ours (Inertial, AIS)} & \textit{\utdmhadimuresult}\\
\textit{Ours (Fused)} & \textit{\utdmhadfuseraw} \\
\textit{Ours (Fused, AIS)} & \textit{\utdmhadfuseresult}\\
% Ours (ASS) & TODO\\
% line\_5\_seperated\_axis & Skeleton & 0.883721\\
% line\_5\_seperated\_axis & Inertial & 0.7 \todo{Redo experiment, number lost}\\
% \midrule
% 3d representation\\
% \midrule
% line\_5 Skeleton & 0.755814 \\
% dot\_5 Skeleton  & 0.709302 \\ 
\bottomrule

\end{tabular}
    \caption{Results on UTD-MHAD. Units are in \%}
    \label{tab:results_utd}

\end{table}

% \begin{table*}[!t]
%     \centering
%     \begin{tabular}{l|r|r|r|r|r|r}
%     \textbf{Approach}  & \textbf{Skeleton} & \textbf{IMU} & \textbf{Motion Capture} & \textbf{Object Estimates} & \textbf{Wifi} & \textbf{Optical Flow}\\%& CS (Top 5)& CV (Top 5)\\
    
%     \toprule
%     Ours      & \cmark & \cmark & \cmark & \cmark & \cmark & \xmark\\
%     Chen \andothers\cite{chen2015utd} & \cmark & \pmark & \xmark & \xmark & \xmark \\
%     Chen \andothers\cite{chen2015utd} & \cmark & \pmark & \xmark & \xmark & \xmark \\
%     \midrule
%     \midrule
    
%     \bottomrule
%     \end{tabular}
%     \caption{Variety. Legend: Shown \cmark, Not shown \xmark, Shown with different approaches \pmark}
%     \label{tab:variety}
% \end{table*}

% \begin{figure}[htbp]
%   \centering
%   \includesvg[width=.9\linewidth]{images/metrics_accuracy_utd_mhad.svg}
%   \caption{Accuracy on the UTD-MHAD dataset. Blue (raw) and red (augmented) represent the accuracy on inertial signals. Pink (raw) and mint (augmented) represent the accuracy on skeleton signals. The augmentation increases the accuracy by more than 10\% per modality.}
%   \label{fig:accuracy_utdmhad}
% \end{figure}

\subsubsection{UTD-MHAD} This dataset \cite{chen2015utd} contains 27 actions of 8 individuals performing 4 repetitions each. RGB-D camera, skeleton estimates and inertial measurements are included.
% The dataset consists of multiple data sources, namely an RGB-D camera, IMU data and skeleton estimates gathered by the Kinect SDK. As a benefit over the other datasets, the UTD-MHAD dataset has two different sensor modalities included. The datasets also include inertial measurements from either the hand or the arm during the movement. 
The RGB-D camera is placed frontal to the demonstrating person. The IMU is either attached at the hand or the leg during the movements. A cross-subject protocol is followed as proposed by the authors \cite{chen2015utd}. Half of the subjects are used for training while the other half is used for validation. 
Results are given in \tabname \ref{tab:results_utd}.
% \subsubsection{WIAR \cite{guo2017novel}} This dataset contains activities recorded with WiFi antennas. We include this dataset for supporting that our proposed approach is transferable to sensor data gathered from others, in the computer vision mostly uncommon, data sources.
% We also applied our approach on the WIAR  benchmark for observing how good the approach works on Wifi signals.

\begin{table}[t]
    \centering
    \begin{tabular}{l|r}
        \textbf{Approach}   & \textbf{Accuracy} \\
        \toprule
        % Baseline &\\
        % \midrule
        Wang \andothers \cite{wang2019joint} & 88.13 \\
        % \midrule
        % Proposed Approach &\\
        % \midrule
        Ours (Raw) & \arilresultraw \\
        % Ours (AIS) &   \textbf{96.04} \\
        \textit{Ours (AIS)} &   \textbf{\arilresult} \\ % Epoch 75
        % Ours (ASS) & \\
        \bottomrule
    \end{tabular}
    \caption{Results on ARIL dataset. Units are in \%}
    \label{tab:results_aril}
\end{table}

\begin{table}[t]
    \centering
        \begin{tabular}{l|r}
        \textbf{Approach}     & \textbf{Accuracy} \\
        \toprule
        Ours (Raw)   & \simitateresultraw \\
        Ours (AIS)   & \simitateresult  \\
        % Ours (Raw, Resnet101)   & 0. & 0. \\
        % Ours (AIS, Resnet101)   & 0. & 0. \\
        % Ours (Raw, Resnet50)   &  0.925926 \\
        % Ours (AIS, Resnet50)   & 0.959064  & 0.994152 \\
        % Ours (Raw, Resnet18)   & 0.925926 & 0.984405 \\
        % Ours (AIS, Resnet18)   & 0.966862 & 0.996101  \\
        \bottomrule
        \end{tabular}
    \caption{Results on Simitate. Units are in \%}
    \label{tab:results_simitate}
\end{table}

\subsubsection{ARIL} This dataset \cite{wang2019joint} contains \wifi Channel State Information (CSI) fingerprints. The CSI describes how wireless signals propagate from the transmitter to the receiver \cite{al2019channel}. A
standard IEEE 802.11n \wifi protocol was used to collect 1398 CSI fingerprints for 6 activities.  The data is varying by location. The 6 classes represent hand gestures \textit{hand circle, hand up, hand cross, hand left, hand down, and hand right} targeting the control of smart home devices. For our experiments, we use the same train/test split as was used by the authors of the dataset (1116 train sequences / 278 test sequences). 
Results are given in \tabname \ref{tab:results_aril}.
% Methods for discriminating the signals are not required as the 52 channels of the CSI fingerprints are already defined to not conflict with each other. We can report a Top-5 accuracy of 1.0 also suggesting the closer inspection of the top confusions in order to find parameters or methods to reduce them.

% \begin{table}[]
%     \centering
% \begin{tabular}{l|r}
% \textbf{Approach}     & \textbf{Accuracy} \\
% \toprule
% \bottomrule
% \end{tabular}
%     \caption{Results on Simitate}
%     \label{tab:results_simitate}
% \end{table}

\subsubsection{Simitate \cite{memmesheimer2019simitate}} The Simitate benchmark focuses on robotic imitation learning tasks. Hand and object data are provided from a motion capturing system in 1932 sequences containing 27 classes of different complexity. The individuals execute tasks of different kinds of activities from drawing motions with their hand over to object interactions and more complex activities like ironing. 
This dataset is interesting as we can fuse human and object measurements from the motion capturing system to add context information. Good action recognition capabilities will allow direct application to symbolic imitation approaches. We use a 80/20 train/test split for our experiments. Results are given in \tabname \ref{tab:results_simitate}.

% Further pre-calculated 2D pose estimates are provided. The accompanied dataset consists of 1932 sequences.  The motivation behind using this dataset is, when those actions can be correctly recognized and the robot has a set of actions that it can map the observed actions to we can imitate the observed behaviour by detection using a Resnet 18 architecture for 100 Epochs.

% If not stated otherwise we trained our approaches with a 80 training / 20 test split. 

\subsection{Results}

We did our best to include results from the most recent approaches for comparison. We found that the proposed representation on a signal level archived good performances across different modalities. An improvement of \arilimpro\% over the baseline has been achieved on a \wifi CSI fingerprint-based dataset \cite{wang2019joint}. Augmentation has shown a positive impact on the resulting accuracy across modalities. The resulting model based on an EfficientNet-B2 performs well in interpreting spatial relations on the color encoded signals across the experiments.
%NTU
For the NTU RGB+D 120 dataset we give results in \tabname \ref{tab:results_ntu}. Related results are taken from literature \cite{liu2019ntu, caetano2019skelemotion, papadopoulos2019vertex}.  A skeleton with 25 joints serves as input for the training of our model. In case multiple identities are contained they are fused with the presented signal fusion approach. 
We got a cross-subject accuracy of \ouresultntucs\% and a cross-view accuracy of \ouresultntucv\% without investment of dataset-specific model tuning.
Intuitively, when considering sequential data, LSTM based approaches are considered. We highly outperform the LSTM based approaches \cite{shahroudy2016ntu,liu2016spatio,liu2017global,liu2017skeleton}. More directly comparable are CNN based approaches \cite{liu2017enhanced,ke2018learning,liu2018recognizing,caetano2019skeleton,caetano2019skelemotion}. All of the mentioned approaches concentrate on finding representations limited to skeleton or human pose features while our approach considers action recognition on a signal level and therefore is transferable to other modalities as well. The discriminative representation we suggest comes closest to the one by Liu \andothers \cite{liu2017enhanced}. In combination with the proposed augmentation method and the EfficientNet-B2 based architecture, we outperform the current CNN based approaches by \ntuimpro. Very recently Papadopoulos \andothers \cite{papadopoulos2019vertex} presented an approach based on a graph convolutional network and performs 5.7\% better on the cross-subject split and 8\% better on a cross-view split than our approach. However, this approach is also limited to graphs constructed from skeleton sequences.
Graph convolutional networks could be an interesting candidate for experiments on multiple modalities in the future. 

% Close to the baseline are our results on the NTU RGB+D 120 datset \cite{liu2019ntu} (-3.44 \%) and UTD-MHAD \cite{chen2015utd} (-0.71 \%) datasets while still benefiting for being generalizable across different low dimensional sensor modalities and without dataset specific model tuning. 

%UTD
Results on the UTD-MHAD dataset are shown in \tabname \ref{tab:results_utd}. We compare our approach to the baseline of the authors as well as a more recent approach \cite{zhao2019bayesian, wang2018action}. While Zhao \andothers \cite{zhao2019bayesian} perform better than our proposed approach we get slightly better results then Wang \andothers \cite{wang2018action} and further have the benefit of being applicable on other sensor modalities. It is to note that the perfect accuracy of 100.0\% in \cite{zhang2019comprehensive} was falsely reported on a similar named dataset. Fused experiments are executed by fusing skeleton estimates and inertial measurements  $\mathbf{S}_{fused} = (\mathbf{S}_{skeleton}|\mathbf{S}_{inertial})$. 
We improve the UTD-MHAD inertial baseline \cite{chen2015utd} by \utdimuimpro\% and the UTD-MHAD skeleton \cite{zhao2019bayesian} baseline by \utdskeletonimpro\%.
The proposed augmentation improved results by +\utdmhadskeletonaugmentimpro\% for Skeletons, by +\utdmhadinertialaugmentimpro\% for IMU data and +\utdmhadfusedaugmentimpro\% for the fusion with the proposed augmentation methods. Fusion in our experiments did now have an overall positive effect. The inertial measurements seem to negatively bias the predicted action. Additional sensor confidence encoding could guide future research.
% \textcolor{red}{
% A Top-5 accuracy of 98.37 should give minor optimization possibilities by inspecting the confusing classes in more detail.  We further show the influence of image-based data augmentation of skeleton and inertial signals in \figname \ref{fig:accuracy_utdmhad} on 100 epochs of training using the 50 layer deep Resnet architecture. For additional result tables and confusion matrices we refer to the attached supplementary material.}
%ARIL
The experiments we conducted on the ARIL dataset are compared to a 1D-ResNet CNN \cite{wang2019joint} architecture proposed by the datasets authors. Results are presented in \tabname \ref{tab:results_aril}. Our approach performs better by \arilimproraw\% and the additional proposed augmentation methods improved the baseline by \arilimpro\%. \wifi CSI fingerprints have the benefit of being separated by their 52 bands already. Signal reduction is therefore not necessary. The additional proposed augmentation methods increase the accuracy by another \arilaugmentimpro\%.

On the Simitate dataset a high accuracy is achieved on an 80/20 train/test split. Results are given in \tabname \ref{tab:results_simitate}. Augmentation on this dataset yields only a minimal improvement.  This dataset is especially interesting for adding context. In addition to the hand poses the object poses can be added by our proposed signal fusion approach. As of now, there are no comparable results published. But the results suggest applicability for 
symbolic imitation approaches in the future.

Most approaches focus on getting high accuracy on a single modality, whereas our approach on a signal level serves as an interesting framework for multi-modal action recognition. In total, we have shown good results across 4 modalities (Skeleton, IMU, MoCap, \wifi). To the authors knowledge, no experiment with a similar extend is known. A huge benefit is the common representation that allows immediate prototyping. Run times are constant, even when additional context or sensors are added due to the representation level fusion. The EfficientNet-B2 architecture serves as a good basis for action recognition on our representation. Additional augmentation has improved the accuracy across the conducted experiments.

%% file: 05_conclusion.tex
% We want to also give a word on ethics as presented approach can be applied on WIFI data that can
% be for passively analyzing activities of persons without them knowing. 
% We presented a representation on a signal level encoding for action recognition. We have shown that our proposed method generalizes across different sensor modalities (Skeleton, Inertial, \wifi) and yield good results in comparison to the baselines on the NTU RGB+D 120 \cite{liu2019ntu} and the UTD-MHAD \cite{chen2015utd} dataset. Further we yield state of the art results on the ARIL \cite{wang2019joint} dataset. All that while still generalizing across other modalities. A signal level encoding for action recognition therefore shows to be a good foundation for multi modal action recognition. The proposed approach also has some interesting properties as additional context or modalities can be reduced to a simple matrix concatenation under the condition that both matrices have a common sequence length. This can be achieved by i.e. sampling.
% A promising dataset \cite{kong2019mmact} focusing on multi-modal and cross-modal action recognition was just recently released to short in advance to integrate it. However it would be a great candidate to verification the proposed approach.
% We hope to foster the research direction of verifying the applicability of proposed approaches across sensor domains.

We propose to transform individual signals of different sensor modalities and represent them as an image. The resulting images are then classified using a EfficientNet-B2 architecture.
Our approach was evaluated on action recognition datasets based on skeleton estimates, inertial measurements, motion capturing data and \wifi CSI fingerprints. 
This is in contrast to many previously proposed approaches that often focus on action recognition on a single modality.
For skeleton data we represent each joint and their respective axis as individual signals. For \wifi we use each of the 52 CSI fingerprint channels as signals. For inertial measurement units we use each axis of the acceleration and angular velocity. For our motion capturing experiments we used each axis of the marker attached to the hand and the interacting objects.
Additional context like subjects and object estimates or even the fusion of different modalities can be flexibly added by a matrix concatenation.
As our approach is limited to sparse signals, we propose filtering methods on a signal level to reduce signals that do not contribute much to the action. By this, additional information can be added without overloading the image representation.
We evaluated our approach on four different datasets. The NTU 120 dataset for skeleton data, the UTD-MHAD dataset for skeleton and inertial data, the ARIL dataset for \wifi data and the Simitate dataset for motion capturing data.
Experimental results show that our approach is achieving good results across the different sensor modalities.